\documentclass[]{article}
\pdfoutput=1
\usepackage{threeparttable}
\usepackage{times}
\usepackage{epsfig}
\usepackage{epstopdf}
\usepackage{amsmath,amsfonts,amssymb,graphicx}
\usepackage{multirow}
\usepackage{colortbl}
\usepackage{hhline}
\usepackage{color}
\usepackage{colortbl}
\definecolor{Gray}{gray}{0.85}
\usepackage{rotating}
\usepackage[nolist, nohyperlinks]{acronym}
\usepackage{fix2col}
\usepackage{pdflscape}
\usepackage{array}
\usepackage{tabularx}
\usepackage{setspace}
\usepackage{hyperref}
\usepackage{lineno}
\modulolinenumbers[5]
\usepackage{longtable}
\usepackage{pdfpages}
\usepackage{array}
\usepackage{authblk}
\renewcommand{\mathbf }{\boldsymbol}
\newcommand{\ignore}[1]{}
\def\eg{e.g.\ }

\def\ie{i.e.\ }

\begin{document} 
\begin{acronym}
\acro{HAR}{Human Action Recognition}
\acro{SoKP}{Sequence of Key Poses}
\acro{BoKP}{Bag of Key Poses}
\acro{DTW}{Dynamic Time Warping}
\acro{DMW}{Dynamic Manifold Warping}
\acro{CTW}{Canonical Time Warping}
\acro{ACA}{Aligned Cluster Analysis}
\acro{HACA}{Hierarchical Aligned Cluster Analysis}
\acro{EA}{Evolutionary Algorithm}
\acro{TPID}{Two Person Interaction Dataset}
\acro{CMU}{Carnegie Mellon University}
\acro{TUM}{Technische Universit\"{a}t M\"{u}nchen}
\acro{MoCap}{Motion Capture}
\acro{MSR}{Microsoft Research}
\acro{IMU}{Inertial Measurement Unit}
\acro{BIT}{Beijing Institute of Technology}
\acro{CAVIAR}{Context Aware Vision using Image-based Active Recognition}
\acro{HOG}{Histogram of Oriented Gradients}
\acro{HOF}{Histograms of Optical Flow}
\acro{STIP}{Space-Time Interest Point}
\acro{SVM}{Support Vector Machine}
\acro{SIFT}{Scale Invariant Feature Transform}
\acro{INRIA}{Institut National de Recherche en Informatique et en Automatique}
\acro{IXMAS}{INRIA Xmas Motion Acquisition Sequences}
\acro{JPL}{Jet Propulsion Laboratory}
\acro{MuHAVi}{Multicamera Human Action Video}
\acro{AoDL}{Activities of Daily Living}
\acro{UMPM}{Utrecht Multi-Person Benchmark}
\acro{MHAD}{Multimodal Human Action Database}
\end{acronym}
\title{From Pose to Activity: Surveying Datasets and Introducing CONVERSE}
\author{Michael Edwards}
\author{Jingjing Deng}
\author{Xianghua Xie}
\affil{Department of Computer Science\\Swansea University\\Faraday Tower, Singleton Park\\Swansea, SA2 8PP\\United Kingdom\\Email: x.xie@swansea.ac.uk}

\vspace*{\fill}
\noindent{}The following paper is a summarized pre-print version of an accepted publication, and therefore portions of the full article have been removed.\newline

For further details we direct the reader to the full paper:
\begin{itemize}
\item[] M. Edwards, J. Deng, X. Xie, \emph{``From Pose to Activity: Surveying Datasets and Introducing CONVERSE''}, Computer Vision and Image Understanding: Special Issue on Individual and Group Activities in Video Event Analysis, 2015.
\end{itemize}
  
\noindent{}The CONVERSE dataset is located at \url{http://csvision.swan.ac.uk/converse}.
\vspace*{\fill}

\maketitle
\begin{abstract}
We present a review on the current state of publicly available datasets within the human action recognition community; highlighting the revival of pose based methods and recent progress of understanding person-person interaction modeling. We categorize datasets regarding several key properties for usage as a benchmark dataset; including the number of class labels, ground truths provided, and application domain they occupy. We also consider the level of abstraction of each dataset; grouping those that present actions, interactions and higher level semantic activities. The survey identifies key appearance and pose based datasets, noting a tendency for simplistic, emphasized, or scripted action classes that are often readily definable by a stable collection of sub-action gestures. There is a clear lack of datasets that provide closely related actions, those that are not implicitly identified via a series of poses and gestures, but rather a dynamic set of interactions. We therefore propose a novel dataset that represents complex conversational interactions between two individuals via 3D pose. 8 pairwise interactions describing 7 separate conversation based scenarios were collected using two Kinect depth sensors. The intention is to provide events that are constructed from numerous primitive actions, interactions and motions, over a period of time; providing a set of subtle action classes that are more representative of the real world, and a challenge to currently developed recognition methodologies. We believe this is among one of the first datasets devoted to conversational interaction classification using 3D pose features and the attributed papers show this task is indeed possible. The full dataset is made publicly available to the research community at \cite{data_CONVERSE}.
\end{abstract}

\section{Discussion of Current State}
\label{sec:Discussion}
Numerous \ac{HAR} datasets have been produced and publicly released in the last decade for the purpose of detecting and identifying action events in an observed scene. Many of these sets have the added benefit of allowing cross-verification of methodologies developed in the field of computer vision; specifically those of action detection and classification. Available datasets contain a variety of traits which require consideration when deciding upon their appropriate usage. Sets differ in the data collection modality; including RGB videos, depth maps, accelerometers and marker based motion capture. They also differ in the actions carried out; including simple gestures, discrete actions, and continuous sequences of actions, multi-user interactions and person-object interactions. Some datasets make use of original data collection, allowing a degree of control over certain parameters within the data collection methodologies. Others use meta-data collected from video clips that are publicly available from media such as films and online video clips; these tend to have large amounts of variation between individual sequences, however they are also among the largest of the datasets, with some meta-sets containing thousands of sequences \cite{Kuehne2011,Soomro2012}. Numerous sets have ground truth labels for an entire sequence; however many are either manually segmented out of a continuous sequence of multiple actions, or are left for users to perform labeling before their use. Ground truth labeling on a frame-by-frame basis is rare, due to the complexity in determining the exact frame at which an action begins. 

Datasets, such as KTH, Weizmann and MSR Action3D \cite{Schuldt2004,Gorelick2007,Li2010}, provide the common examples of well annotated and discrete action executions; including kicking, walking, and shaking. Others, such as the CMU Motion Capture set \cite{data_CMUMoCap}, expand the complexity further by containing sequences of multiple actions executed in a continuous manner. Recently, sets have moved towards recognizing interaction between two people, including SBU Kinect, BIT-Interaction and K3HI \cite{Yun2012,Kong2012,Hu2013}; however, these sets still provide interactions using the classic simplistic actions of pushing, punching and kicking. A few studies, including MSR DailyActivity3D and the TUM Kitchen \cite{Wang2012,Tenorth2009}, have made steps towards the recognition of so-called 'daily activities', natural actions which may be more representative of the real world executions. 

Despite this abundance of datasets, there is still a lack of sets that make use of subtle interaction classes, representing loosely defined actions such as those in natural conversational styles, or in context dependent situations. With \cite{Deng2013k,Deng2013}, we have presented methodology, using a dataset of subtle conversational interactions, which is able to classify such subtle action events, based upon 3D pose features.

{\singlespacing
\begin{table*}[!t]

\begin{scriptsize}
\begin{center}
\caption{Comparisons of key action recognition datasets, detailing the download location, associated descriptive publications, and number of simultaneous viewpoints.}
\begin{tabularx}{\textwidth}{llm{1.5cm}m{1.5cm}m{1.5cm}}
\\\hline
Name & Modality & URL & Description & Views\\\hline
50 Salads & RGB-D, IMU & \cite{data_50Salads} & \cite{Stein2013}  & 1\\
BEHAVE & RGB & \cite{data_BEHAVE} & \cite{Blunsden2010}& 2\\
Berkeley MHAD & RGB-D, IMU, Audio, MoCap & \cite{data_BerkeleyMHAD} & \cite{Ofli2013} & 14\\
BIT Interaction & RGB & \cite{data_BITInteraction} & \cite{Kong2012} & 1\\
CAD120 & RGB-D & \cite{data_CAD60120} & \cite{Sung2012} & 2\\
CAD60 & RGB-D & \cite{data_CAD60120} & \cite{Koppula2013} & 2\\
CASIA & RGB & \cite{data_CASIA} & \cite{Huang2009b}  & 3\\
CAVIAR & RGB & \cite{data_CAVIAR} & \cite{Fisher2004} & 1, 2\\
CMU MMAC & RGB, MoCap, IMU & \cite{data_CMUMMAC} & \cite{DelaTorre2008} & 6\\
CMU MoCap & MoCap & \cite{data_CMUMoCap} & - & 1\\
\bf{CONVERSE} & RGB-D & \cite{data_CONVERSE} &\cite{Deng2013a,Deng2013k,Deng2013} & 1\\
Drinking/Smoking & RGB & \cite{data_DrinkSmoke} & \cite{Laptev2007} & 1\\
ETISEO & RGB & \cite{data_ETISEO} & \cite{Nghiem2007} & 1, 3, 4\\
G3D & RGB-D  & \cite{data_G3D} & \cite{Bloom2012} & 1\\
G3Di & RGB-D  & \cite{data_G3Di} & \cite{Bloom2014} & 1\\
HMDB51 & RGB & \cite{data_HMDB} & \cite{Kuehne2011} & 1\\
Hollywood & RGB & \cite{data_Hollywood} & \cite{Laptev2008} & 1\\
Hollywood-2 & RGB & \cite{data_Hollywood2} & \cite{Marszalek2009} & 1 \\
Hollywood3D & RGB-D & \cite{data_Hollywood3D} & \cite{Hadfield2013} & 1\\
HumanEVA-I & RGB, MoCap & \cite{data_HumanEva} & \cite{Sigal2010} & 7\\
HumanEVA-II & RGB, MoCap & \cite{data_HumanEva} & \cite{Sigal2010} & 4\\
IXMAS & RGB, Silhouette & \cite{data_IXMAS} & \cite{Weinland2006} & 5\\
JPL & RGB & \cite{data_JPL} & \cite{ryoo2013first} & 1\\
K3HI & RGB-D & \cite{data_K3HI} & \cite{Hu2013} & 1\\
KTH & RGB & \cite{data_KTH} & \cite{Schuldt2004} & 1\\
LIRIS & RGB-D & \cite{data_LIRIS} & \cite{Wolf201414} & 1\\
MPI08 & RGB, IMU, Laser Scan & \cite{data_MPI08} & \cite{PonBaa2010a,BaaHel2010a} & 8\\
MPII Cooking & RGB & \cite{data_MPIICook} & \cite{Rohrbach2012Cook} & 1\\
MPII Composite & RGB & \cite{data_MPIIComp} & \cite{Rohrbach2012Comp} & 1\\
MSR Action-I & RGB & \cite{data_MSR} & \cite{Yuan2009} & 1\\
MSR Action-II & RGB & \cite{data_MSR} & \cite{Yuan2011} & 1\\
MSR Action3D & RGB-D & \cite{data_MSR} & \cite{Li2010} & 1\\
MSR DA3D & RGB-D & \cite{data_MSR} & \cite{Wang2012} & 1\\
MSR Gesture3D & RGB-D & \cite{data_MSR} & \cite{Kurakin2012} & 1\\
MuHAVi & RGB, Silhouette & \cite{data_MuHAVi} & \cite{Singh2010} & 8\\
Olympic Sports & RGB & \cite{data_Olympic} & \cite{Niebles2010} & 1\\
POETICON & RGB, MoCap & \cite{data_POETICON} & \cite{Wallraven2011} & 7\\
Rochester AoDL & RGB & \cite{data_RAODL} & \cite{iccv2009MessingPalKautz} & 1\\
SBU Kinect Interaction & RGB-D & \cite{data_TPID} & \cite{kiwon_hau3d12} & 1\\
Stanford 40 Actions & Image & \cite{data_Stanford40} & \cite{Yao2011a} & 1\\
TUM Kitchen & RGB, Markerless MoCap, RFID & \cite{data_TUM} & \cite{Tenorth2009} & 4\\
UCF101 & RGB & \cite{data_UCF101} & \cite{Soomro2012} & 1\\
UCF11 & RGB & \cite{data_UCF11} & \cite{Liu2009} & 1\\
UCF50 & RGB & \cite{data_UCF50} & \cite{Reddy2012} & 1\\
UCF Sport & RGB & \cite{data_UCFSport} & \cite{Rodriguez2008d} & 1\\
UMPM & RGB, MoCap & \cite{data_UMPM} & \cite{HICV11:UMPM} & 1\\
UT Interaction & RGB & \cite{data_UT} & \cite{Ryoo2009d} & 1\\
ViHASi & RGB, Silhouette & \cite{data_ViHASi} & \cite{Ragheb2008} & 40\\
VIRAT & RGB & \cite{data_VIRAT} & \cite{Oh2011} & - \\
Weizmann & RGB, Silhouette & \cite{data_Weizmann} & \cite{ActionsAsSpaceTimeShapes_iccv05,Gorelick2007} & 1\\
WVU MultiView & RGB & \cite{data_WVU} & \cite{Ramagiri2011,Kavi2013} & 8\\\hline
\end{tabularx}
\label{tbl:Cite}
\end{center}
\end{scriptsize}
\end{table*}
}

The following section will evaluate the public datasets detailed within section \textbf{OMITTED FROM SUMMARY} and summarized in Table \ref{tbl:Cite}, identifying key features for their usage in the \ac{HAR} community. Several parameters that require consideration when developing and evaluating action recognition methodologies using publicly available data are identified; including the modality of data acquisition, data provided by the set, and consistent training and testing subsets. The complexity of each dataset is also evaluated, based upon the number of individual classes they present, the number of samples provided, and the presence of complex and realistic class scenarios. Summaries are provided in Tables \ref{tbl:Cite}, \ref{tbl:Data}, \ref{tbl:Action}, \ref{tbl:Size}, \ref{tbl:Domain}, \ref{tbl:Truth}, \ref{tbl:Scene}, and \ref{tbl:Popularity}. The proposed CONVERSE dataset \cite{data_CONVERSE} is included within the evaluations to highlight the necessity for such a set and identify where it resides amongst the currently available data. A detailed explanation of the proposed dataset is given in section \ref{sec:Proposed}.

\subsection{Modality}
\begingroup
\newcolumntype{L}[1]{>{\raggedright\let\newline\\\arraybackslash\hspace{0pt}}m{#1}}
\renewcommand{\arraystretch}{1.5}
\begin{table}[!t]
\begin{center}
\begin{scriptsize}
\begin{threeparttable}
\caption{Comparison of provided data and presence of dedicated validation sets.}
\begin{tabularx}{\textwidth}{lp{4.5cm}p{4.5cm}}
\hline 
& \multicolumn{2}{c}{Datasets} \\\hline
\emph{Data} & & \\
RGB/Greyscale &\multicolumn{2}{l}{All sets except CMU MoCap, K3HI, } \\
MoCap & \multicolumn{2}{L{9cm}}{Berkeley MHAD, CMU MMAC, CMU MoCap, HumanEVA-I, HumanEVA-II, POETICON, TUM Kitchen, UMPM} \\
Depth & \multicolumn{2}{L{9cm}}{50 Salads, Berkeley MHAD, CAD120, CAD60, G3D, G3Di, Hollywood3D, LIRIS, MSR Action3D, MSR DA3D, MSR Gesture3D, SBU Kinect Interaction, \bf{CONVERSE}} \\
Skeleton & \multicolumn{2}{L{9cm}}{Berkeley MHAD, CAD120, CAD60, G3D, G3Di, K3HI, MSR Action3D, MSR DA3D, SBU Kinect Interaction, \bf{CONVERSE}} \\
\ac{IMU}& \multicolumn{2}{L{9cm}}{50 Salads, Berkeley MHAD, CMU MMAC, MPI08, TUM Kitchen} \\
Audio & \multicolumn{2}{L{9cm}}{Berkeley MHAD, POETICON} \\
Laser Scan & \multicolumn{2}{L{9cm}}{MP108} \\\hline 
& \multicolumn{1}{c}{Appearance sets} & \multicolumn{1}{c}{Pose sets}\\\hline
\emph{Train/Test split} & & \\
Yes & Drinking/Smoking, ETISEO, Hollywood, Hollywood 2, IXMAS\footnotemark[1], KTH, Olympic Sports, Rochester AoDL\footnotemark[1], Stanford 40 Actions, UCF101, UCF11\footnotemark[1], UCF50\footnotemark[1], UCF Sport\footnotemark[1], UT Interaction, ViHASi\footnotemark[1], VIRAT\footnotemark[1], Weizmann\footnotemark[1], WVU MultiView-I, WVU MultiView-II & Hollywood3D, HumanEVA-I, HumanEVA-II, LIRIS, MSR Action3D, SBU Kinect Interaction, TUM Kitchen\footnotemark[1], \bf{CONVERSE}\footnotemark[1] \\
No & BEHAVE, BIT-Interaction, CASIA, CAVIAR, HMDB51, JPL, MPII Cooking, MPII Composite, MSR Action-I, MSR Action-I, MuHAVi & 50 Salads, Berkeley MHAD, CAD120, CAD60, CMU MMAC, CMU MoCap, G3D, G3Di, K3HI, MPI08, MSR DA3D, MSR Gesture3D, POETICON, UMPM\\\hline 
\end{tabularx}
\label{tbl:Data}
\begin{tablenotes}
\item[1] provided in description paper via Leave Out cross validation methodology
\end{tablenotes}
\end{threeparttable}
\end{scriptsize}
\end{center}
\end{table}
\endgroup

In Table \ref{tbl:Data} we cluster the datasets based on their method of data capture; from video, depth maps, skeletal tracking, \ac{MoCap} marker tracking, \ac{IMU}, and audio. The majority of sets in \ac{HAR} make use of vision, however recent progress has been made towards the use of 3D pose estimation via depth sensors; therefore understanding the modality provided by a dataset will often impact on the choice of features used to describe each sequence. 

\subsubsection*{Video}
Appearance based \ac{HAR} makes use of datasets that are often collected via still images or video, as cameras can provide a relatively cost effective method of obtaining both real-world and staged execution samples from both a laboratory or real-world environment. In Table \ref{tbl:Cite} it can be seen that all of the datasets presented contain some form of video or appearance based data (except CMU MoCap, K3HI and UCF iPhone), therefore in Table \ref{tbl:Data} we omit the video data. The quality of the recordings varies greatly between sets, with some specializing in evaluating action detection and recognition in low quality or small scale recordings. High intra-set and inter-sequence variation in image quality, camera motion, scale and viewpoint are common in meta-data sets that collect observations from multiple sources, such as UCF101, UCF50, UCF11, Hollywood, Hollywood-2 and HMDB51, and these pose a more realistic problem to the community. Visual based \ac{HAR} can provide an intuitive representation of the scene, however there can often be superfluous information contained within an observation that negatively impacts on the reliable global recognition of a given action; therefore, appearance based modalities can often make use of subject localization and background removal, coupled with the extraction of descriptors such as \ac{STIP}s, \ac{HOG}, \ac{HOF} or local regions of motion features to enable the global recognition of actions regardless of background information or subject-specific appearance. Many depth based datasets also provide simultaneously captured video representations of their data; this appearance data can either be omitted from the learning, or combined to form a multi-modal system. Of the appearance based datasets, the KTH and Weizmann datasets have been cited the most for single action recognition method evaluation. For appearance based interaction recognition the CAVIAR, Hollywood and UT Interaction datasets have been used frequently by the community. 

\subsubsection*{\ac{MoCap}}
Motion capture concerns the recording of numerous markers placed upon the body by multi-camera systems, providing accurate tracking of the markers within a volume over time. \ac{MoCap} often provides a method of capturing a spatial ground truth for the marker locations within the scene, being used as a stand-alone modality or augmenting datasets captured through other methods. \ac{MoCap} systems are often calibrated using built in software and a calibration tool, allowing all cameras to be spatially and temporally synchronized, increasing confidence in the marker tracking. Placement of the markers varies between datasets and as such datasets which make use of \ac{MoCap} provide details of the marker placement on the body, allowing semantic affordance to be applied to each marker. \ac{MoCap} can be seen as a cost-expensive method of data collection, often requiring dedicated systems, however the generation of a spacial ground truth and reliable pose tracking method is of great benefit when developing pose from appearance or pose based action recognition methodologies. Despite this, an implementation of marker based \ac{MoCap} systems in a real world environment is impractical, requiring individuals to wear a motion capture suit to be detected by the system would provide little benefit to the user; as such there has been some effort has also been made to produce human skeletal tracking without the use of markers from simple RGB image recording \cite{Tenorth2009} and from depth maps \cite{Li2010}. 

Of the \ac{HAR} datasets that utilize MoCap, the HumanEVA, Berkeley \ac{MHAD} and \ac{CMU} \ac{MoCap} datasets are most commonly used. The HumanEVA dataset provides a set of evaluation metrics for the purpose of action recognition, Berkeley \ac{MHAD} provides a detailed dataset containing multiple modalities for fusion based action recognition, and the \ac{CMU} \ac{MoCap} dataset contains a vast number of continuous sequences which can be used for action detection and sequence segmentation. 

\subsubsection*{Depth}
The production of a consumer level depth sensor, most notably the Microsoft Kinect, coupled with efficient and accurate joint tracking software has provided the \ac{HAR} community with an inexpensive method of collecting 3D poses of a subject performing actions within a scene \cite{Andersen2012, Berger2013, Han2013}. This has allowed for the development of methods that represent the action as a series of key poses or bag of words model \cite{Parameswaran2003,Chaaraoui2012,Deng2013}, extracting the key frames that describe the overall action event. Datasets such as 50 Salads, Berkeley \ac{MHAD}, CAD120, CAD60, G3D, G3Di, K3HI, LIRIS, MSR Action3D, MSR DA3D, MSR Gesture3D, and SBU Kinect Interaction all make use of the Kinect depth sensor to collect data providing the depth map of the scene. The Hollywood3D set utilizes commercial films that have been recorded using a 3D stereo camera system to provide depth maps. By obtaining a 3D pose estimation of the subjects within the scene users are able to, given accurate tracking, generate pose, scale, and appearance invariant features for the purpose of \ac{HAR} that include joint trajectories, joint-joint distances, joint-plane distances, and joint motion histories. Many of the depth datasets captured using the Kinect provide the associated estimated skeleton representation of the individual, tracking a number of joints across the scene. The number of joints tracked and the position of the provided markers often depends on the method used to extract the skeleton; those using the Microsoft Kinect SDK often provide 20 points, whilst those using the OpenNI standard track 15 joints on the body. The selection of joints often aligns with the major joints of the human body, and so provides an estimation of limb motion. Currently the use of depth sensors are limited to a viewpoint that is in a roughly front-on position due to the method of estimating depth, using distortions of infra-red projections into the scene which is then captured by a receiving sensor. This method has little ability to handle scene occlusions which can cause shadowed regions in the depth map, resulting in lost or noisy tracking in the extracted skeletons.   

The most prominent depth datasets for single person actions include those presented by the Microsoft Research group, namely the Action3D and DA3D datasets. Despite the small number of samples and action classes provided by the MSR Action3D dataset there has been a vast number of citations for its use as an evaluation dataset. For person-person interactions there are few datasets available which make use of depth based data; the K3HI and SBU Kinect Interaction datasets provide sequences of single executions of a given interaction, analogous to those provided by the BIT Interaction and UT Interaction appearance datasets, however their recent release may reflect their low citation and usage for evaluation of pose based methods. 

\subsubsection*{Other}
Various other methods of data capture have been used for \ac{HAR} purposes, including the use of audio recordings \cite{Stork2012,Ofli2013} and \ac{IMU}s \cite{Ermes2008, McCall2012, Ofli2013}. These methods can provide reasonable classification results on their own, however they are often used in a multi-modality system to improve the accuracy rates of single modality methods. These datasets are beyond the scope of this survey and omitted for brevity.

\subsection{Action class types}
\begin{figure}
\begin{center}
\begin{scriptsize}
\includegraphics[scale = 0.5]{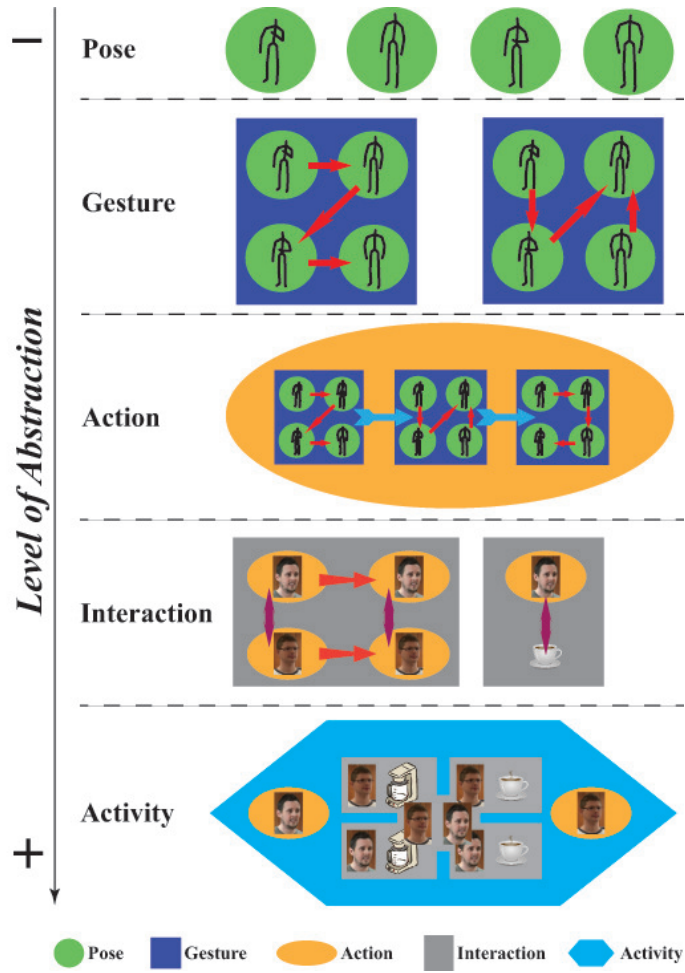}
\caption{Levels of abstraction within human action recognition.}
\label{fig:abstraction}
\end{scriptsize}
\end{center}
\end{figure}

Human behaviors are often a set of events with differing levels of abstraction and complexity, therefore to aid comparison between \ac{HAR} class types we shall first define assumptions made about terminology we wish to use. Many class labels provided within \ac{HAR} datasets can often be re-labeled to fit within a different level of abstraction, however we attempt to use common terminology found across the community, with an overview provided in Figure \ref{fig:abstraction} and a summary of the datasets in Table \ref{tbl:Action}. Example images from datasets that describe differing levels of abstraction are given in Table \textbf{OMITTED FROM SUMMARY}.
\newcommand{\picwidth}{2cm}
\newcommand{\piclength}{2.5cm}
\newcommand{\datasetwidth}{0.8cm}
\newcommand{\typewidth}{0.5cm}

\begin{description}
\item[Pose] An atomic observation of the spatial arrangement of a human body at a single temporal instance, \eg `Arm above head'.
\item[Gesture] A temporal series of poses on a sub-action scale, sometimes described as action primitives \eg `Arm moves left'.
\item[Action] A series of gestures which form a contextual event, \eg Repeated gestures of arm moving left and then right can be contextual described as an 'overhead wave action'. These are the most commonly used class labels found within current datasets, describing single actions executed by a subject including `run',`jump', and `wave'. 
\item[Interaction] A pairwise or reciprocal action is committed by two entities on each other. Each entity therefore has a single action that reflects it's state compared to the other entity, \ie consider the action of person A shaking the hand of person B; A executes the action of shaking the hand of B, B executes the action of having their hand shaken by A, together this pairwise action execution can be described as that of a 'handshake' interaction. For the purpose of action recognition interactions are often further divided into differing interaction types based on if the entities include people, objects or groups. For this study we have omitted group interaction datasets due to space limitations.
\begin{description}
\item[Person-Person] An action is committed directly by one individual upon another. This definition does not include crowded scenes in which an individual performs a single person action with other subjects in the environment. The class labels in a P-P interaction treats the interaction as a single entity, rather than two separate single person actions, \eg we consider the class `punching' as an interaction between person A, the puncher, and person B, the individual being punched.
\item[Person-Object] An action is committed directly by one individual upon an object. This includes the manipulation of objects. We consider class labels such as `lift chair' and `open box' as person-object interactions as the actions `lift' and `open' are performed on the objects `chair' and `box' respectively.
\item[Groups] Characterized as interactions carried out between a collected entity of more than two individuals. Group interactions can include inter- and intra-group behaviors and the interaction of the group on other objects, individuals, or even other groups. These often form their own subsets of group behaviors.
\end{description}
\item[Activity] A collection of actions and/or interactions that compound to describe a high level event. These are common within the sets that describe daily behaviors, \eg `cook a meal' and `tidy room' can often include numerous actions and interactions that are executed. Each action and interaction can therefore be thought of a sub-activity event in such scenarios. Activity is also used to describe the daily activities, a more realistic observation execution than the exaggerated instances such as 'punch' and 'kick'.
\end{description}

A common scenario presented within \ac{HAR} instances is that of a single person executing a singular action, in which an individual actor performs an action with no interaction to other individuals or objects, such as within KTH, Weizmann, MSR Action, and MSR Action3D. In recent years, interaction datasets have become more prominent, often displaying actions where one actor performs an action upon which another actor is the recipient. These interaction sets can still exhibit behaviors that are quite well defined, with a single instigator and a single recipient, such as punching, pushing and move towards. The most notable interaction sets include BIT Interaction, UT Interaction, K3HI, and SBU Kinect Interact datasets. There also exists interaction classes that are more complex in their composition, involving multiple entities, object manipulation or requiring higher level semantics; these are prominent in the TUM, BEHAVE, VIRAT, ETISEO, and POETICON datasets. The higher level activity datasets often provide observations of an entire task being carried out and require the understanding of the sub-activity actions and interactions being carried out over the course of the recording. In the current sets there are often annotations of lower level actions which are encompassed within a higher level activity context, with sets such as MPII Composite, 50 Salads and TUM Kitchen providing annotations of both levels of abstraction and the objects that are subject to interactions during the course of the activity.

The choice of classes that are performed by the actors is a key motivation in the generation and usage of the proposed dataset. Often the actions executed are those of a visually definable nature, comprising single executions of a discrete action which contain key poses and gestures. The complexity of the problem can then be increased by observing multiple executions of actions in a sequence, either with distinct boundaries between the classes or with a natural flow between different classes. These are all complex issues that are the focus of the community, with segmentation methods often utilized to separate out actions from a continuous sequence. Judging the difference in complexity between two classes can be subjective, depending upon the subtlety of gestures, the context of any interactions, and the spatio-temporal rigidity of the executions; subtle gestures, for example, may well present a more complex recognition problem than the simplest of activity classes. We can however make some generalized assumptions about the complexity within the different abstraction levels. Lower levels of abstraction such as pose and gesture should provide less challenges to the field in its current state, while higher levels of abstraction, especially those involving interactions between two or more entities, still remain a challenging issue. 

Obviously with the definitions of the action types presented there can be some overlap in how to handle events in which an entity is not only interacted with, but also pivotal to the context of the label. Consider the class label `smoking', this event can fit both into the definition of a singular action in which the object is explicit to the action, a person-object interaction between the person and cigarette, and also into its own activity class in which smoking is the task executed. Consider also the class label of `pushing', this may be a class label that can be readily classified as a single action, person-person interaction, or person-object interaction depending upon the entities present, and also as an activity if there is a contextual background to the event. This highlights the complexity in describing class labels and requires the careful consideration of overlaps that appear to be presented between datasets with similar action classes. To further this point, we ask should the community consider an interaction as its own complete class, or should the system understand the states occupied by all entities within the interaction, \ie the class label of `pushing' may be deconstructed into sub-classes that describe the action of the instigator and the reaction of the recipient. Many interaction datasets handle the class labeling as a single complete unit of interaction, often reliant on the action committed by the instigator, \eg K3HI, SBU Kinect Interaction, and UT Interaction. However the TUM Kitchen, 50 Salads and MPII Composite sets explicitly annotate the states of both entities to define the person-object interactions for the purpose of activity recognition. The use of a single interaction class that encompasses all sub-divisions of that interaction may provide learning that is broad and resistant to variation of intra-class behaviors; however by learning the sub-divisions of an interaction class, considering the different actions and reactions as their own states, there may be an ability to learn more effective boundaries for execution variations. For this study we have considered and evaluated upon the class labels provided by the original datasets; however we invite the community towards potentially defining multi-scale class labeling for the purpose of action and activity recognition.

\begingroup
\renewcommand{\arraystretch}{1.5}
\begin{table}[!t]
\begin{center}
\begin{scriptsize}
\begin{threeparttable}
\caption{Comparison of dataset interaction types. Note that datasets can contain instances of several types of behaviors based on the labeling it provides.}
\begin{tabularx}{\textwidth}{lp{4cm}p{4cm}}
\hline
& \multicolumn{1}{c}{Appearance sets} & \multicolumn{1}{c}{Pose sets}\\ \hline
\emph{Event type} & &  \\
Action & CASIA, CAVIAR, Drinking/Smoking, ETISEO, HMDB51, Hollywood, Hollywood-2, IXMAS, KTH, MSR Action-I, MSR Action-II, MuHAVi, UCF11, UCF Sports, ViHASi, VIRAT, Weizmann, WVU MultiView-I, WVU MultiView-II  & 50 Salads, Berkeley MHAD, CAD120, CAD60, CMU MoCap, G3D, Hollywood3D, HumanEVA-I, HumanEVA-II, LIRIS, MPI08, MSR Action3D, MSR Gesture3D, POETICON, TUM Kitchen, UMPM\\
Interaction: Person - Person & BEHAVE, BIT Interaction, CASIA, CAVIAR, ETISEO, Hollywood, Hollywood-2, JPL, UT Interaction & CMU MoCap, G3Di, Hollywood3D, K3HI, LIRIS, POETICON, SBU Kinect Interaction, UMPM, \bf{CONVERSE} \\
Interaction: Person - Object & ETISEO, MPII Cooking, MPII Composite, VIRAT & 50 Salads, CAD120, CMU MMAC, LIRIS, POETICON, TUM Kitchen, UMPM\\
Activity & CASIA, MPII Composite, MuHAVi, Olympic Sports, Rochester AoDL, Stanford 40 Actions, UCF101, UCF11, UCF50, UCF Sports, ViHASi & 50 Salads, MSR DA3D, CAD60, LIRIS, TUM Kitchen, \bf{CONVERSE} \\\hline
\end{tabularx}
\label{tbl:Action}
\begin{tablenotes}
\item
\end{tablenotes}
\end{threeparttable}
\end{scriptsize}
\end{center}
\end{table}
\endgroup

\subsection{Size}
The size of a dataset, not just in the number of sequences but also in the range of different action classes and participants, can impact on it's suitability for method evaluation. Testing on a small-scale dataset can provide misleading results during analysis which may not be replicated when introducing more class labels or observations, due in part to the highly variable nature of inter- and intra-instance executions. Contrarily there are implications in the usage of large datasets; not only the collection and storage of data, but also in the  processing of features, class learning and validation. Due to the inherent issues in obtaining a large number of participants, action classes, and sequences, the largest sets tend to be meta-sets, which collect action sequences from various sources, such as YouTube and films, containing large variation between sequences; this often makes meta-sets highly variable and challenging problems to be solved. A summary of dataset sizes is given in Table \ref{tbl:Size}

\subsubsection*{Number of classes}
Datasets with a small number of action classes, such as MSR Action-I, MSR Action-II, and Drinking/Smoking, can often provide strong recognition results in part due to the low number of partitions needed to divide the actions provided within the set. Those sets that contain a large number of action classes, namely HMDB51, UCF101, and UCF50, provide a difficult challenge to \ac{HAR} methods due to the need to find partitioning information within each class that allows for inter-class partitioning, whilst preserving intra-class similarity. Due to the inconceivable number of possible actions and interactions that can exist in the real world it can be beneficial to evaluate methodologies on datasets with a large number of distinct action classes. 
\subsubsection*{Number of subjects}
Datasets that are able to provide more individual subjects performing an action are able to portray the variability in inter- and intra-subject execution of a given class. Observations of the same action class can often differ greatly in both their temporal rate and spatial occupancy, leading to complexity in learning the action for recognition purposes. Methods that are able to provide subject invariant action recognition should provide consistent results on a dataset which contains a large number of subjects. Again, the meta-sets tend to provide the highest number of subjects, almost capturing a new subject per sequence, representing a large range of inter-subject variation.
\subsubsection*{Number of samples per class}
The number of observations per class can impact on the ability of a system to suitable learn a given class. A low number of observed instances of a class can result in weak recognition of unobserved instances of the same class. HMDB51 provides over 100 instances of each action class it contains, providing a range of observations across differing viewpoints, quality and executions, as such it can provide a useful benchmark for the recognition of actions from a subject and observation invariant methodology. Current pose based datasets contain few repeated instances of an action class, often with 3-5 repetitions per subject per class. To increase the number of instances per class it is possible to segment those datasets which contain continuous recordings of multiple executions into discrete single execution clips, this includes the KTH dataset.
\subsubsection*{Number of sequences}
The total number of sequences within a dataset should be a factor of the number of subjects, classes, and number of class executions, and as such can impact on the reliability of the results produced. Larger datasets can provide larger testing sets for which to evaluate a system, allowing for more confidence in the results of the validation. Size alone however is only one parameter in the selection of evaluation benchmark, with domain, class complexity and modality impacting on the application of methodologies to real world implementations.

\begingroup
\renewcommand{\arraystretch}{1.5}
\begin{table}[!t]
\begin{center}
\begin{scriptsize}
\begin{threeparttable}
\caption{Comparison of dataset sizes.}
\begin{tabularx}{\textwidth}{lp{4.8cm}p{4.8cm}}
\hline
& \multicolumn{1}{c}{Appearance sets} & \multicolumn{1}{c}{Pose sets}\\ \hline
\emph{\# Actions}&& \\
$\leq$ 5 & Drinking/Smoking, MSR Action-I, MSR Action-II  & \\
6 - 10 &  BEHAVE, BIT Interaction, CAVIAR, Hollywood, Hollywood-2, JPL, KTH, Rochester AoDL, UCF Sport, UT Interaction, Weizmann, WVU MultiView-II & CMU MMAC, HumanEva-I, HumanEva-II, K3HI, LIRIS, MPI08, POETICON, SBU Kinect Interaction, UMPM, \bf{CONVERSE}\\
11 - 15 & CASIA, ETISEO, IXMAS, UCF11, VIRAT, WVU MultiView-I & Berkeley MHAD, CAD60, G3Di, Hollywood3D, MSR Gesture3D, TUM Kitchen\\
16 - 20 & MuHAVi, Olympic Sports, ViHASi & 50 Salads, CAD120, G3D, MSR Action3D, MSR DA3D\\
$\geq$ 21	& HMDB51, MPII Cooking, MPII Composite, Stanford 40 Actions, UCF101, UCF50 & CMU MoCap\\\hline
\emph{\# Subjects} & & \\
$\leq$ 5	&Rochester AoDL & CAD120, CAD60, HumanEVA-I, HumanEVA-II, MPI08, POETICON, TUM Kitchen\\[.05cm]
6 - 10 &MSR Action-I, MSR Action-II, UT Interaction, ViHASi, Weizmann & G3D, MSR Action3D, MSR DA3D, MSR Gesture3D, SBU Kinect Interaction\\
11 - 20 & IXMAS, MPII Cooking, MuHAVi & Berkeley MHAD, G3Di, K3HI, \bf{CONVERSE}\\
$\geq$ 21 & CASIA, KTH, MPII Composite & 50 Salads, CMU MMAC, CMU MoCap, UMPM\\
Undefined & BEHAVE, BIT Interaction, CAVIAR, Drinking/Smoking, ETISEO, HMDB51, Hollywood, Hollywood-2, JPL, Olympic Sports, Stanford 40 Actions, UCF101, UCF11, UCF50, UCF Sport, VIRAT, WVU MultiView-I, WVU MultiView-II & Hollywood3D, LIRIS \\\hline 
\emph{\# Sequences} & & \\
$\leq$ 20 & BEHAVE, CAVIAR, MSR Action-I,UT Interaction, WVU MultiView-II & HumanEVA-II, TUM Kitchen, \bf{CONVERSE} \\
21 - 100 & ETISEO, JPL, MPII Cooking, MSR Action-II, Weizmann & 50 Salads, CAD60, CMU MMAC, G3Di, HumanEVA-I, MPI08, POETICON, UMPM\\
101 - 500 & BIT Interaction, Drinking/Smoking, Hollywood, MPII Composite, Rochester AoDL, UCF Sport, ViHASi & CAD120, G3D, K3HI, MSR DA3D, MSR Gesture3D, SBU Kinect Interaction \\
501 - 1000 & KTH, Olympic Sports, WVU MultiView-I & Berkeley MHAD, Hollywood3D, LIRIS, MSR Action3D \\
$\geq$ 1001 & CASIA, Hollywood2, HMDB51, IXMAS, MuHAVi, Stanford 40 Actions, UCF101, UCF11, UCF50, VIRAT & CMU MoCap \\\hline 
\end{tabularx}
\label{tbl:Size}
\begin{tablenotes}
\item
\end{tablenotes}
\end{threeparttable}
\end{scriptsize}
\end{center}
\end{table}
\endgroup

\subsection{Application Domain}
The intended application domain of a dataset can provide certain intrinsic features in the data collection methodology and action classes captured, from low resolution images of CCTV surveillance footage to more complex action sequences of daily living. Some actions are representative of the domain from which they are intended; for example the UCF-Sports dataset, \cite{Rodriguez2008d}, makes use of numerous actions from various sports, such as javelin throws and long jumps. We classify the datasets into 4 action class domains; generic actions, daily living, surveillance, and sport. Generic action datasets have no overall theme, instead providing classes that are pan-domain; these include the classes `running', `jumping', `punching', and also more complex interactions such as `handshake' or `play guitar'. Daily living datasets often include actions and activities that are more natural in their execution and environment, this includes classes based on assisted living and household tasks. Surveillance datasets often make use of elevated view points and lower resolution images, mirroring the common camera setups in the security industry \cite{Ferryman,Oh2011}. Sports based action recognition often makes use of previously captured data from multiple sources, often containing varying image quality and varying levels of camera motion. A summary of the domains for each of the datasets is provided in Table \ref{tbl:Domain}.

\subsubsection*{Generic}
Many action recognition datasets often contain generic action classes that are observable in numerous domains. The intention is to cover a wide variety of actions to allow domain invariant action recognition, with generic datasets being the most widely used for validation purposes, including the KTH \cite{Schuldt2004}, Weizmann \cite{Gorelick2007} and MSR Action3D \cite{Wang2012} sets. Many generic datasets are collected in a laboratory environment; with static cameras, static backgrounds and calibrated data-capture setups, including Berkeley MHAD and CMU MoCap. Others may be collected outdoors with a controlled clutter free setting, such as Weizmann and KTH. Others are collected within cluttered environments, featuring non-participatory subjects that complicate the scene, such as MSR Action-I and Action-II. Pose based datasets which make use of a depth sensor and the pose estimation technique of extracting the 3D skeleton are often captured in a relatively clutter free scene due to the limitations of the skeletal tracking methodology used.

\subsubsection*{Daily living}
Daily living sets are designed to closely represent the natural world in both the environmental surroundings and the natural style of action classes executed. The Tum Kitchen \cite{Tenorth2009}, MSR DA3D \cite{Li2010,Wang2012}, MPII Cooking \cite{Rohrbach2012}, and Rochester AoDL \cite{Messing2009} sets are commonly used for the analysis of methodology in the recognition of day-to-day activities. Activities include `having a conversation', `phone calls', `laying down', `drinking' and `eating', but may also include sub-actions within a higher level task, such as `setting a table' or `cooking a meal'. The executions may be allowed to occur naturally as in the 50 Salads, MPII Cooking, and MPII Composite datasets; or the observations may be more scripted, such as in the POETICON and the robotic class of the TUM Kitchen set \cite{Wallraven2011,Stein2013, Tenorth2009}. By understanding the actions and interactions within a daily activity dataset the field is moving towards learning higher level semantics of human behavior via natural representations.

\subsubsection*{Surveillance}
Surveillance is a domain concerned with detecting and identifying activity within a continuous observation of a scene, often making use of video-based action recognition samples that are taken from a distance, prone to crowding, and contain poor resolution recordings. A surveillance domain sequence may contain more frames of empty or redundant information, sporadically interspersed with temporally short regions of interest. Datasets such as UT-Interaction, CASIA, and BEHAVE make use of surveillance style setups to capture emphasized person-person interaction classes such as `come together' and `fight'. The CAVIAR, ETISEO, and VIRAT datasets all make use of detailed ground truth annotations to provide information regarding persons and objects within the scene, enabling the evaluation of methods in detecting varies entities and their interactions within a scene for higher semantic understanding of the events.

\subsubsection*{Sport}
The UCF-Sports, \cite{Rodriguez2008d}, and Olympic Sports, \cite{Niebles2010}, datasets are focused explicitly on sports related action examples. These sets contain samples that are collected from various sources of TV and online recordings, providing samples that vary in their recording quality and containing both static and dynamic camera movements. As such these can often be challenging datasets. In both cases the intent of the dataset is to be able to recognize the sport being performed, this can be more challenging than in the case of learning sports related actions, such as in the case of `tennis serve' and `boxing' from some of the generic action datasets. A sport as a high level class can contain numerous action and interaction actions that make up the overall activity and learning a sporting class may require learning vastly different observations that belong to the same class. 3D pose based \ac{HAR} in the sports domain has few datasets due to the complexity in capturing a large volume in which the activity can be played. The G3Di dataset provides interactions between two people in the context of a sporting game played through a console, however we treat the provided classes as being generic actions rather than true sporting based actions. 

\begingroup
\renewcommand{\arraystretch}{1.5}
\begin{table}[!t]
\begin{center}
\begin{scriptsize}
\begin{threeparttable}
\caption{Comparison of dataset domain applications.}
\begin{tabularx}{\textwidth}{lp{4cm}p{4cm}}
\hline
& \multicolumn{1}{c}{Appearance sets} & \multicolumn{1}{c}{Pose sets}\\ \hline
\emph{Domain} & &  \\
Generic&BIT Interaction, HMDB51, Hollywood, Hollywood-2, IXMAS, JPL, KTH, MSR Action-I, MSR Action-II, MuHAVi, Stanford 40 Actions, UCF101, UCF50, UCF11, ViHASi, Weizmann, WVU MultiView & Berkeley MHAD, CMU MoCap, G3D, G3Di, Hollywood3D, HumanEVA, K3HI, MPI08, MSR Action3D, MSR Gesture3D, SBU Kinect Interaction, UMPM\\
Daily Living& Drinking/Smoking, MPII Cooking, MPII Composite, Rochester AoDL & 50 Salads, CAD120, CAD60, CMU MMAC, LIRIS, MSR DA3D, POETICON, TUM Kitchen, \bf{CONVERSE}\\
Surveillance& BEHAVE, CASIA, CAVIAR, ETISEO, UT-Interaction, VIRAT &\\
Sport & Olympic Sports, UCF Sports & \\\hline
\end{tabularx}
\label{tbl:Domain}
\begin{tablenotes}
\item
\end{tablenotes}
\end{threeparttable}
\end{scriptsize}
\end{center}
\end{table}
\endgroup

\subsection{Ground truth}
Table \ref{tbl:Truth} outlines various ground truths provided with each dataset, both for spatial ground truths and labeling of action classes. Providing consistent ground truth with which to evaluate results is important for developing benchmarks against which to test developed methodologies, aiding in the generation of a metric score that can be used to compare implementations.  

Class label ground truths and scene annotations of a dataset can provide a clear benchmark for quantifying the performance of a developed methodology. Some datasets provide frame-by-frame labeling of the scene, whilst others label an entire sequence as containing a given class label. These annotations allow quantification of results obtained from various methodologies, with predicted class labels and detections being compared against the ground truth. The collection of the class ground truth can be either manually annotated by the author or produced via some form of machine learning. Manual annotation can provide detailed descriptions of the entire scene, with locations and affordances being given to persons and objects within the scene, as can be seen with the ETISEO and HMDB51 datasets. These can be extremely useful when tracking the states of multiple entities within the scene, or for the understanding of a high level abstracted class; however the manual labeling of individual frames can produce observation bias into the dataset, requiring strict objective criterion to gain consistent ground truths. Machine based annotations can combined machine learning with data labeling to rapidly provide ground truths to large datasets, \eg the Hollywood and Hollywood-2 datasets are partially annotated by learning textual descriptions within the film's scripts. An automated ground truth annotation may require subsequent manual verification to ensure the false labeling is minimized. The simplest form of ground truth labeling provided by \ac{HAR} datasets is by attributing the entire sequence to a specific label, acknowledging that a given action occurs at some point within the observation, as is the case with CASIA, CMU MMAC, MSR Action3D, and many more. Having simplistic whole sequence labeling can make it hard to use such datasets for detection purposes, as evaluating the beginning and end frames of an action can be problematic to determine manually. For action recognition purposes the learning of background frames from a sequence may also provide some level of noise to the partitioning of that class. 

Spatial truth can be provided by explicitly locating the subjects and objects within the environment or by highlighting regions of interest in which the the subject, object or event resides by using bounding boxes or silhouette masks. Calibrated ground truth methods can be used to determine the spatial locations of the subjects within a scene, often using motion capture suits and markers to explicitly track the body through a capture volume, providing either a raw point cloud or the predicted skeletal frame of the body. The accuracy of motion capture systems can vary from method to method, however the resolution accuracy is often within a range of a few millimeters, providing superior body tracking than using machine learning based pose extraction. Marker based motion capture systems, such as those used in CMU \ac{MoCap} and Berkeley MHAD, require the application of each marker to the individual at certain predetermined locations, and variation in placement of the markers on the body from sequence to sequence can introduce small errors in obtaining truly explicit spatial truths. The use of depth maps to extract an estimated 3D pose of the subject in the scene has become a prominent inclusion in depth based \ac{HAR} datasets such as MSR Action3D, K3HI, SBU Kinect Interaction, CAD120, and CAD60. The observation is fed into a skeleton extractor, such as the OpenNI, Microsoft Kinect SDK softwares, or custom methods \cite{Girshick2011,Shen2012,Taylor2012}, in which a subject is located and a human skeleton model is fitted, predicting the 3D coordinates for a number of joints. Although an approximation of true 3D spatial orientation of the joints, depth sensors and joint tracking has been shown to be relatively accurate in the tracking of humans \cite{Andersen2012, Han2013}.
The use of bounding boxes to describe regions of interest in a scene are common within appearance based datasets, such as BEHAVE, CAVIAR, ETISEO and MSR Action, especially those that consider person-object interactions or belong to the surveillance domain. They simply provide an area of focus that contains relevant annotated information, such as object and subject location. The use of silhouette masks also provide a region of interest, whilst simultaneously removing external and internal appearance information, representing the subject as a binary classification as either belonging to the background or foreground. These regions of interest can also be utilized to validate action detection and localization methodologies, removing the unwanted information from the overall observation.

\begin{table*}[!t]
\begin{scriptsize}
\begin{center}
\caption{Description of ground truths provided by datasets.}
\begin{tabularx}{\textwidth}{llp{5.5cm}}
\\\hline
Name & Spatial ground truth labels & Class ground truth labels\\\hline
50 Salads & - &Frame labeling\\
BEHAVE & Bounding boxes & Frame annotation\\
Berkley MHAD & MoCap tracking & File labeling\\
BIT Interaction & -& File labeling\\
CAD120 & Extracted skeleton, bounding boxes& Frame labeling\\
CAD60 & Extracted skeleton& File labeling\\
CASIA & -& File labeling\\
CAVIAR & Bounding box & Frame labeling\\
CMU MMAC & MoCap tracking & File labeling\\
CMU MoCap & MoCap tracking& File labeling\\
\bf{CONVERSE} & Extracted skeleton & Frame labeling\\
Drinking/Smoking & Bounding box & Frame labeling\\
ETISEO & Bounding box & Frame labeling including calibration parameters, scene descriptions, object affordance\\
G3D &Extracted skeleton& File labeling\\
G3Di &Extracted skeleton& File labeling\\
HMDB51 & Bounding boxes& File labeling including view, camera motion, visible body parts, quality, and number of subjects\\
Hollywood &- & Frame labeling\\
Hollywood-2 &- & Frame labeling\\
Hollywood 3D &- & File labeling\\
HumanEVA-I & MoCap tracking & File labeling\\
HumanEVA-II & MoCap tracking & File labeling\\
IXMAS & Silhouette masks & Frame labeling\\
JPL & - & Frame labeling\\
K3HI & Extracted skeleton & File labeling\\
KTH & - & Frame labeling including scenario labeling\\
LIRIS & Bounding boxes & Frame labeling\\
MPI08 & MoCap tracking and 3D scan & File labeling\\
MPII Cooking & - & Frame labeling\\
MPII Composite & - & Frame labeling\\
MSR Action-I & Bounding box & Frame labeling\\
MSR Action-II & Bounding box & Frame labeling\\
MSR Action3D & Extracted skeleton & File labeling\\
MSR DA3D & Extracted skeleton & File labeling\\
MSR Gesture3D & Extracted skeleton & File labeling\\
MuHAVi & Silhouette masks & Frame labeling\\
Olympic Sports & - & File labeling\\
POETICON & MoCap tracking& File labeling\\
Rochester AoDL & - & File labeling\\
SBU Kinect Interaction & Extracted skeleton & File labeling\\
Stanford 40 Actions & Bounding box & File labeling\\
TUM Kitchen & Markerless MoCap tracking & Frame labeling including body trunk, left arm, right arm, and object affordance\\
UCF101 & - & Frame labeling\\
UCF11 & - & Frame labeling\\
UCF50 & - & Frame labeling\\
UCF Sport & - & File labeling\\
UMPM & MoCap tracking & File labeling\\
UT Interaction & Bounding box & Frame labeling\\
ViHASi & Silhouette masks & File labeling \\
VIRAT & Bounding box & Frame labeling including object affordance\\
Weizmann & Silhouette masks & File labeling\\
WVU MultiView-I & - & File labeling\\
WVU MultiView-II & - & File labeling\\\hline
\end{tabularx}
\label{tbl:Truth}
\end{center}
\end{scriptsize}
\end{table*}

\subsection{Viewpoint}
Camera based methods can also make use of various viewpoints, from single camera to multi-camera simultaneous viewpoint capture. Viewpoints can also differ greatly, capturing events from roughly a parallel plane with the ground, elevated above head height, or from an almost top-down viewpoint. Often events are captured from a viewpoint that is roughly parallel to the ground, producing observations that are almost representative of a human-eye view of the event, examples can be found in MSR Action3D, K3HI, and CMU MoCap. A summary of dataset viewpoint representation is given in Table \ref{tbl:Scene}. Sets such as BEHAVE, UT Interaction and CASIA contain events recorded from an elevated angle; these viewpoints are common within the surveillance domain due to the positioning of surveillance cameras for capturing a large scene at once.  Recently there has been work towards the recognition of actions from a first person perspective, with data captured from the viewpoint of the observer \cite{ryoo2013first,FathiFirstPerson,RyooRoboCentric}. This field is often working towards the understand of interactions by robots for the purpose of human-robot interaction. Such a viewpoint is believed to provide more meaningful information when the observer has an active role in the interaction rather than simply observing a scene, as is the case in human-robotics interactions. There are also datasets which attempt to capture simultaneous multi-camera views of an event for the purpose of evaluating supposedly pose-invariant methodologies. Sets such as WVU MultiView, Berkeley MHAD and TUM Kitchen all contain numerous cameras located in differing positions capturing the same scene. Depth based data, such as tracked skeletons and motion capture marker coordinates, can be orientated arbitrarily about its three axes to develop multi-view methodology, with some pose alignment used to reduce the effect of orientation discrepancies, \cite{Chaaraoui2014a}. However this is dependent upon accurate pose estimation in order to provide data which has confident tracking. Due to the nature of extracting pose estimation from depth based methods there are limited numbers of datasets that utilize multiple depth sensors; however Berkeley MHAD provides multiple Kinect recordings alongside it's vast number of appearance views, with the sensors located in positions from which the infrared sensors are not causing occlusions. 

\begingroup
\renewcommand{\arraystretch}{1.5}
\begin{table}[!t]
\begin{center}
\begin{scriptsize}
\begin{threeparttable}
\caption{Comparison of dataset viewpoints and scenario control.}
\begin{tabularx}{\textwidth}{lp{4.5cm}p{4.5cm}}
\hline
& \multicolumn{1}{c}{Appearance sets} & \multicolumn{1}{c}{Pose sets}\\ \hline
\emph{Simultaneous Views} & &  \\
Monocular & BIT Interaction, Drinking/Smoking, HMDB51, Hollywood, Hollywood-2, JPL, KTH, MPII Cooking, MPII Composite, MSR Action-I, MSR Action-II, Olympic Sports, Rochester AoDL, Stanford 40 Actions, UCF101, UCF11, UCF50, UCF Sport, UT Interaction, Weizmann & 50 Salads, CMU MoCap, G3D, G3Di, Hollywood3D, K3HI, LIRIS, MSR Action3D, MSR DA3D, MSR Gesture3D, SBU Kinect, UMPM\\
Multi-view & BEHAVE, CASIA, CAVIAR, ETISEO, IXMAS, MuHAVi, TUM Kitchen, ViHASi, WVU MultiView-I, WVU MultiView-II & Berkeley MHAD, CAD120, CAD60, CMU MMAC, HumanEVA-I, HumanEVA-II, MPI08, POETICON, \bf{CONVERSE}\\\hline 
\emph{Environment} & &  \\
Interior Natural & CAVIAR, Drinking/Smoking, HMDB51, Hollywood, Hollywood-2, JPL, MuHAVi, Olympic Sports, Stanford 40 Actions, UCF101, UCF11, UCF50 & Hollywood3D \\
Interior Controlled & IXMAS,MPII Cooking, MPII Composite, Rochester AoDL, ViHASi, WVU MultiView-I, WVU MultiView-II & 50 Salads, Berkeley MHAD, CAD120, CAD60, CMU MMAC, CMU MoCap, G3D, G3Di, HumanEva-I, HumanEva-II, K3HI, LIRIS, MPI08, MSR DA3D, MSR Gesture3D, POETICON, SBU Kinect Interaction, TUM Kitchen, UMPM, \bf{CONVERSE}\\
Exterior Natural & BEHAVE, BIT Interaction, Drinking/Smoking, ETISEO, HMDB51, Hollywood, Hollywood-2, MSR Action-I, MSR Action-II, Olympic Sports, Stanford 40 Actions, UCF101, UCF11, UCF50, UT Interaction, VIRAT & Hollywood3D  \\
Exterior Controlled & BIT Interaction, KTH, Weizmann & \\\hline                       
\end{tabularx}
\label{tbl:Scene}
\begin{tablenotes}
\item
\end{tablenotes}
\end{threeparttable}
\end{scriptsize}
\end{center}
\end{table}
\endgroup

\subsection{Use in Community}
Popularity of a dataset within the community can be difficult to evaluate, however here we attempt to identify the number of citations that are made to the dataset's description publication via Google Scholar. Using this count as a measure of how well adopted a given dataset has become, we rank each set in Table \ref{tbl:Popularity}. Note that older sets can often show higher citation due in part to their steady accumulation of references over time. Similarly, the number of citations made may not explicitly reflect the use of dataset as a benchmark, as often the datasets are published in parallel with a novel methodology which may accrue its own citations. It can be seen from Table \ref{tbl:Popularity} that the pose based datasets show considerably fewer citations, most likely due to the relative age of the rapidly growing field.

{\singlespacing
\begin{table*}[!t]

\begin{scriptsize}
\begin{center}
\caption{Citation count for dataset description paper. Correct at time of submission. Note: CMU MoCap has no attributed publication}
\begin{tabularx}{7cm}{lcc}
\\\hline
Name & Year of Publication & Total Citations\\\hline
\textbf{\underline{Appearance}} & &\\
KTH & 2004 & 2013\\
Hollywood & 2008 & 1772\\
Weizmann & 2005 & 1182\\
UCF11 & 2009 & 602\\
IXMAS & 2006 & 590\\
UCF Sport & 2008 & 584\\
Hollywood-2 & 2009 & 580\\
Drinking/Smoking & 2007 & 327\\
UT Interaction & 2009 & 303\\
Olympic Sports & 2010 & 283\\
Rochester AoDL & 2009 & 266\\
HMDB51 & 2011 & 265\\
MSR Action-I & 2009 & 189\\
UCF101 & 2012 & 155\\
VIRAT & 2011 & 144\\
Stanford 40 Actions & 2011 & 137\\
UCF50 & 2013 & 131\\
ETISEO & 2007 & 103\\
CAVIAR & 2004 & 90\\
MSR Action-II & 2011 & 82\\
MPII Cooking & 2012 & 67\\
MuHAVi & 2010 & 60\\
MPI08 & 2010 & 48\\
JPL & 2013 & 38\\
ViHASi & 2008 & 33\\
BEHAVE & 2010 & 33\\
MPII Composite & 2012 & 32\\
BIT Interaction & 2012 & 19\\
CASIA & 2009 & 12\\
WVU MultiView & 2011 & 0\\\hline

\textbf{\underline{Pose}} &&\\
HumanEVA & 2010 & 373\\
MSR Action3D & 2010 & 333\\
MSR DA3D & 2012 & 311\\
CAD120 & 2012 & 159\\
TUM Kitchen & 2009 & 117\\
CAD60 & 2013 & 81\\
MSR Gesture3D & 2012 & 75\\
Berkeley MHAD & 2013 & 50\\
CMU MMAC & 2008 & 48\\
SBU Kinect Interaction & 2012 & 33\\
Hollywood3D & 2013 & 32\\
G3D & 2012 & 28\\
POETICON & 2011 & 8\\
UMPM & 2011 & 7\\
50 Salads & 2013 & 6\\
LIRIS & 2014 & 5\\
\bf{CONVERSE} & 2015 & 4\\
K3HI & 2013 & 2\\
G3Di & 2014 & 0\\
CMU MoCap & - & -\\\hline
\end{tabularx}
\label{tbl:Popularity}
\end{center}
\end{scriptsize}
\end{table*}
}

\section{Proposed Dataset}
\label{sec:Proposed}
In the following section we draw upon the findings from the survey to present our own novel dataset for the recognition of complex conversational interactions between two individuals. We outline the necessity for the production of the set, the structure of the dataset and report on several previous publications that have utilized the dataset.
\subsection{Requirement for the Dataset}
As can be seen from the previous sections, datasets that are able to capture human action using appearance based modalities, such as RGB videos, have developed from representing non-realistic emphasized actions to considering more complex interactions between individuals and their surrounding environment. The field has moved from actions which are easily distinguishable in the visual domain, \eg `waving' and `jumping', to those of interactions, although still recognizable, \eg `hug' and `kiss' \cite{Dollar2005d,MI_LEARN}. Due to the availability of these datasets many methods have been produced and evaluated for the purpose of action recognition and detection, including the use of \ac{SIFT} \cite{Scovanner2007}, temporal Harris corner features \cite{Laptev2005} or \ac{STIP}s \cite{Laptev2008}.

Meanwhile the depth based methodology which has risen to prominence over the past decade has far fewer publicly available datasets which consider the problem of person-person interactions, with most considering either emphasized actions or interactions. As such we believe that the publication of a dataset that represents highly complex person-person interactions is timely. We have chosen to capture conversational interactions between two individuals using the Kinect depth sensor, posing the challenge of recognizing subtle interaction classes.

The primitive action provided by many of the available datasets can be decomposed into a series of definable gestures and atomic poses. However we argue that real-world social interactions contain more complex and subtle class partitioning, being a product of multiple actions, semantics and the interplay between those involved. We therefore propose the problem of recognizing interactions in which the distinguishing features are containing within the temporal dynamics of the total event, such as that of a verbal interaction. We provide a dataset in which the interaction is labeled as a whole, rather than describing the event based on the primitive gestures within the scene. By providing such a dataset we hope to move the field towards the recognition of scenarios in which the defining descriptors are highly complex and context specific. 
\subsection{Apparatus Setup}
\begin{figure}
\begin{center}
\begin{scriptsize}
\includegraphics[scale = 0.4]{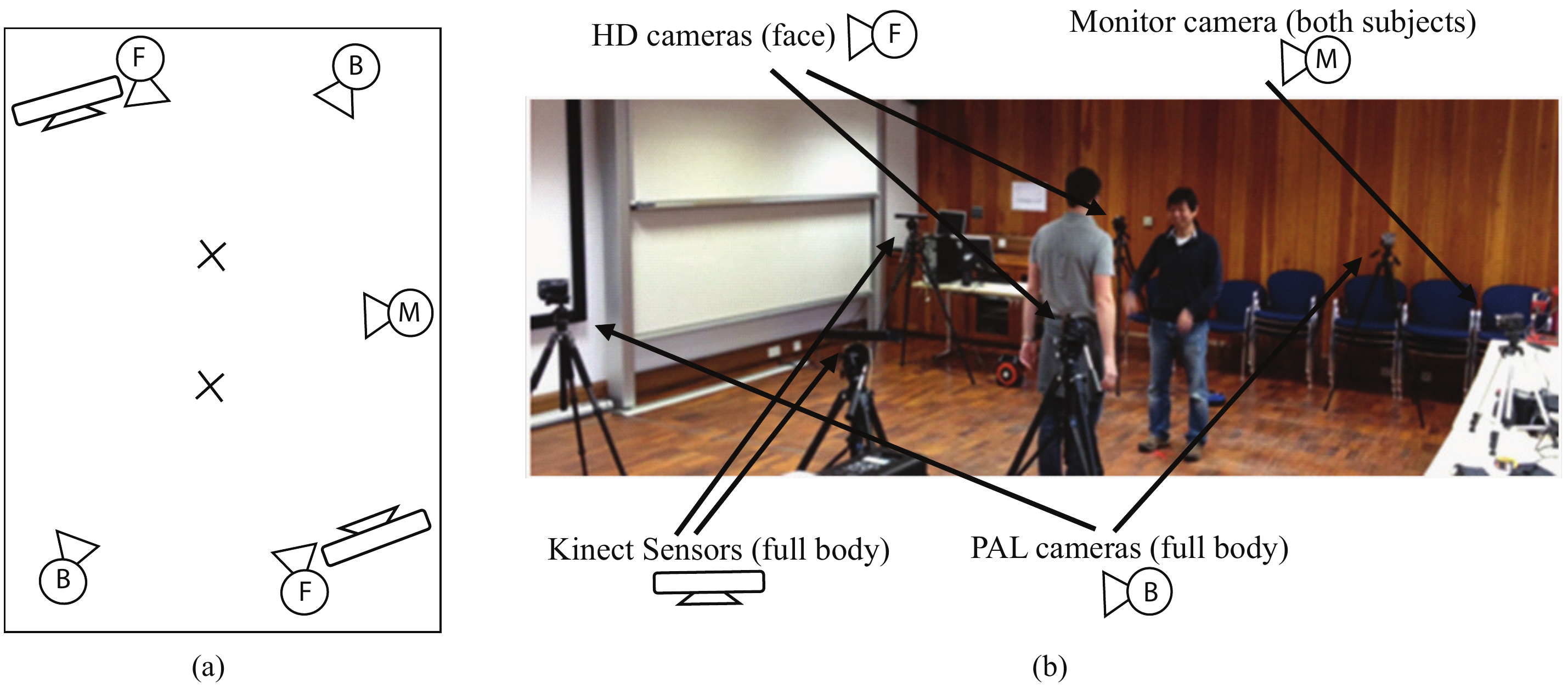}
\caption{Layout of the CONVERSE data capture environment. a) Plan view of the capture environment b) Photo showing subjects within the cluttered environment.}
\label{fig:setup}
\end{scriptsize}
\end{center}
\end{figure}

In this work, we choose seven conversational action categories and use a two-Kinect setup to capture 3D human pose during the interaction between two individuals. The collection environment consisted of a cleared space within a boardroom (Figure \ref{fig:setup}); in order to keep the dataset complex, no effort was made to homogenize the environment by use of any backdrops. Two Kinect sensors were located at opposite ends of the room, approximately two meters away from a marker on which a subject would be loosely located. Each person was recorded using a single Kinect Sensor at 30fps. The Kinect was offset to the front right of the subject in order to avoid occlusion from the opposing subject, which could occur if taking a frontal recording of the subject. Subjects were placed approximately one meter apart but not limited in their movement. Two PAL cameras (B cameras in Figure \ref{fig:setup}) were located to capture the full body of a single participant, with a third camera (M in Figure \ref{fig:setup}) located to capture the entire recording scene. These recordings are purely for the monitoring of the experiment and synchronization, thus are not provided within the dataset published in \cite{data_CONVERSE}. Cameras were also located to capture the face of each participant (F cameras in Figure \ref{fig:setup}), these provide the RGB recordings used to generate the gaze estimation provided. The recording devices were not located in the same place, and as such there is orientation variance between the depth maps and the RGB recordings.

\subsection{Action Descriptions}
Participants were required to complete 7 different conversational tasks, outlined in Table \ref{table:TaskDescription}. There was no time limitation on the execution of each task, and some tasks took naturally longer than others. Several tasks were given revealed to the participants before collection, to allow preparation; the actions that required preparation were describing work, story telling, debate, discussion and jokes. If the participants were given the problem or subjective question before the study then there may have been a reduction in interaction between the individuals. 

Each task was performed and then there was a small break while the participants were reminded of the next task to carry out. The first task was to discuss an area of their current work. The second task was to prepare an interesting story to tell their partner, such as a holiday experience. The third task was to jointly find the answer to a problem. The fourth task was a debate, where the participants were asked to prepare arguments from opposing view points on an issue we gave to them. In the fifth task they were asked to discuss the issues surrounding a particular statement and come to agreement whether they believe the statement is true or not. The participants were asked to reach an agreement through discussion; hence, it is different to the debate task, which was based on conflicting views. The sixth task was to answer a subjective question, and the seventh task was to take it in turn telling jokes to one another.

\begingroup
\renewcommand{\arraystretch}{1.5}
\begin{table}[!t]
\begin{center}
\begin{scriptsize}
\begin{threeparttable}
\caption{
Description of each of the tasks given to the participants to perform. The rightmost column describes whether the participants were told about the task and asked to prepare before attending.
} 

\begin{tabularx}{\textwidth}[th]{p{2cm}p{6cm}p{3cm}} 
\hline
Task Name  &  Description & Prepared in advance\\
\hline
Describing Work & Each participant describes their current work or project to partner. The partner then repeats the description back, to confirm they had understood. & Yes\\
Story Telling & Participant were asked to think of an interesting story they could tell their partner. & Yes\\
Problem Solving & Participants were given the problem ``Do candles burn in space and if so what shape and direction?", and asked to think of the solution of together. & No\\
Debate & Participants prepared arguments for a given point of view, pro or con, on the topic ``Should University education be free?", and then debated this between them. & Yes\\
Discussion & Participants were asked to jointly discuss issues surrounding the statement ``Social Networks have made the world a better place", and come to agreement whether they believe the statement is true or not. & No\\
Subjective Question & Participants responded to the subjective question ``If you could be any animal, what animal and why?" & No\\
Telling jokes & Participants were asked to take it in turn telling three separate jokes. & Yes\\
\hline
\end{tabularx}

\label{table:TaskDescription} 
\begin{tablenotes}
\item
\end{tablenotes}
\end{threeparttable}
\end{scriptsize}
\end{center}
\end{table}
\endgroup

\subsection{Participants}
16 subjects responded to a call for participants to take part in dataset collection and provided their consent for the collection. Participants were then organized into 8 pairs to record the person-person interaction during the following series of conversational styles. Interested individuals were asked to prepare for tasks `Describing Work', `Story Telling', `Debate' and `Joke' in advance, while the topics for `Problem Solving', `Discussion' and `Subjective Question' were provided during collection. Participants were not subjected to time limitations or any execution styles. 

\begin{figure}
\begin{center}
\label{fig:PictureBoard}
\includegraphics[scale=0.3]{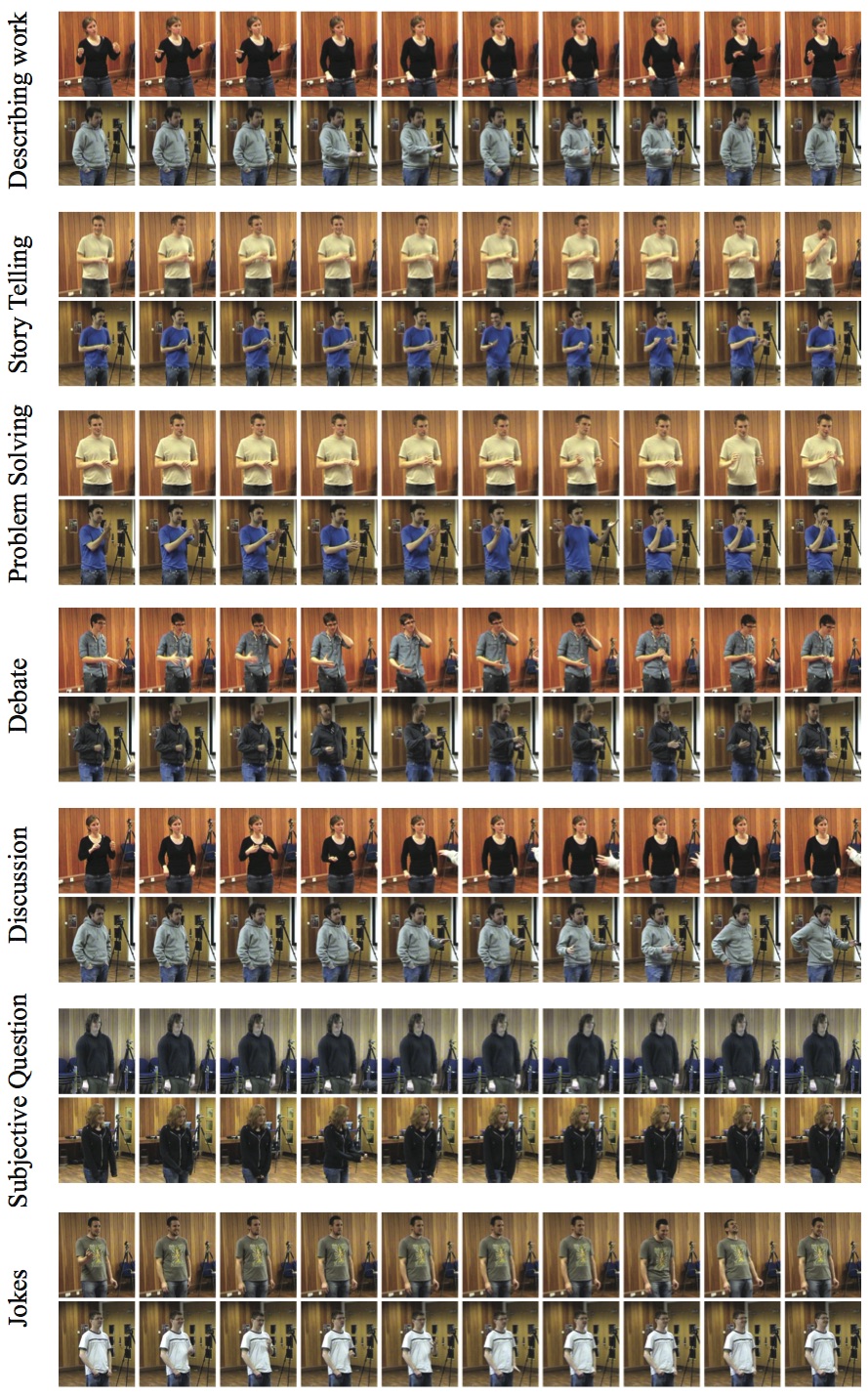}
\caption{Example recordings from each of the 7 action classes, sampled at 2 second intervals and omitting the lower half of the body.}
\end{center}
\end{figure}

\begin{figure}
\begin{center}
\label{fig:SkeletonBoard}
\includegraphics[scale=0.3]{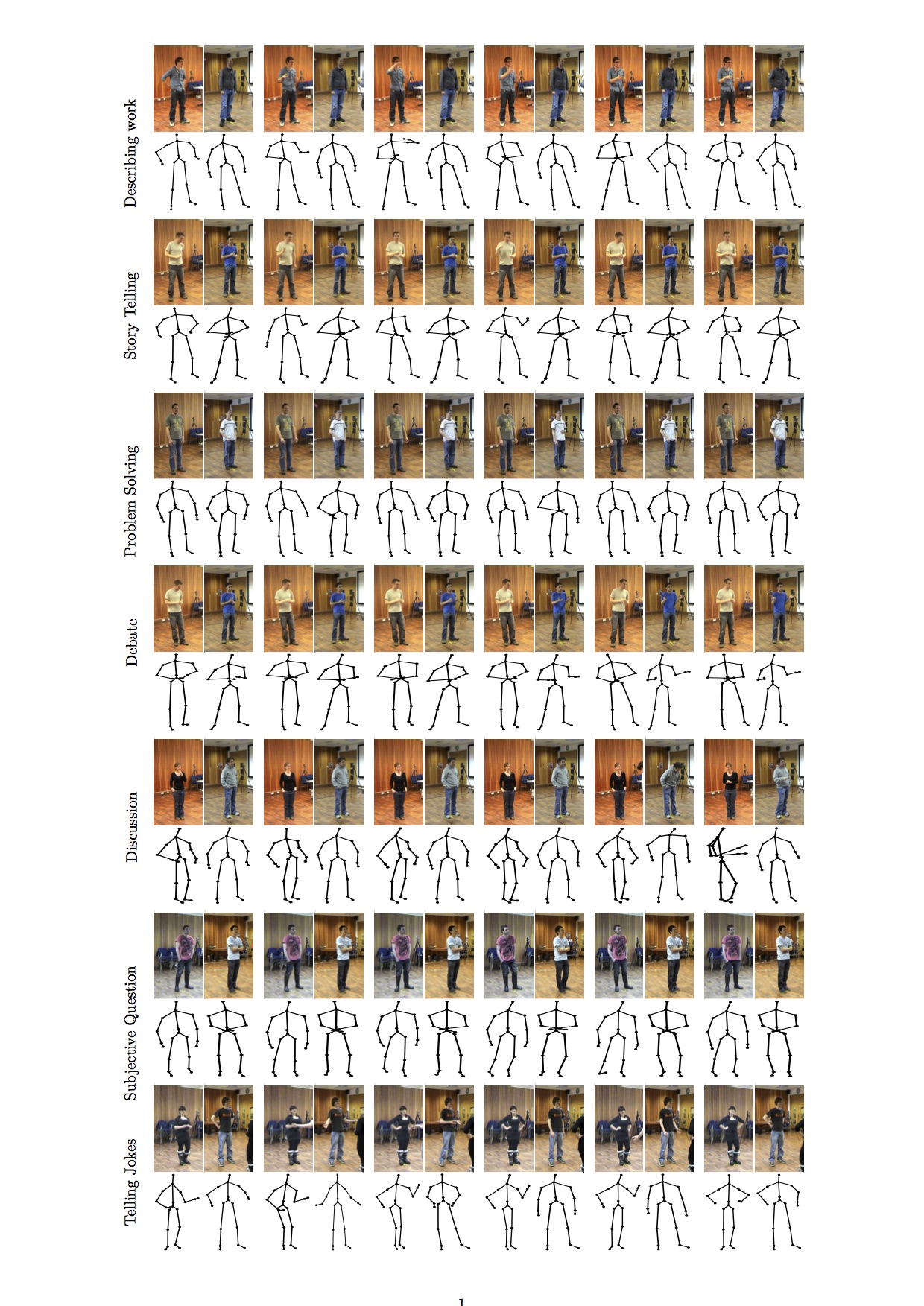}
\caption{Example recordings and skeleton poses from each of the 7 action classes, sampled at 2 second intervals and omitting the lower half of the body.}
\end{center}
\end{figure}

\subsection{Data Provided}
The main data in the collection is the skeletons extracted using the Microsoft Kinect SDK, providing the 20 tracked joints and the confidence of the tracking at each frame in the sequence. The raw depth and RGB recordings from the Kinect are also available alongside the RGB recordings from the separate camcorder. We also provide facial tracking features used for the tracking of gaze and facial dynamics which have been used for feature fusion in \cite{Deng}. Despite the benefit that audio provides to action classification \cite{Peskin93topicand,110003214205,Ververidis20061162,Ofli2013}, the audio has been stripped from all recordings due to the private natures of the conversations that occurred during the interactions. This allows the conversations to be natural, providing a more realistic representation of the scenarios than if each subject was given a script. Although this may be disappointing to those wishing to carry out audio-visual feature fusion, we believe that CONVERSE provides a more complex challenge to be solved when occluding the audio cues of conversation.

\subsection{Results Obtained on CONVERSE}
To provide insight into the use of CONVERSE for interaction recognition we provide baseline results achieved using various state of the art methods for subject-specific classification, with results reported in Tables \ref{table:cmvw} and \ref{table:cmvt}. To achieve this level of accuracy we followed the methods outlined in \cite{Deng}; utilizing pose, face and head orientation features to provide a visual vocabulary of words and topics. Discriminative classifiers, SVM and Random Forest (RF), were trained to classify each CONVERSE task based on the discriminative power of the features. K-Nearest Neighbor (KNN) was selected as a baseline classification technique for comparison. First a Gaussian Mixture Model (GMM) was fitted to low level features (joint-joint/joint-plane distances and joint velocity) in order to obtain a vocabulary of 740 visual words consisting of the Gaussian components taken from 5 second clips, 370 words from facial features and 370 from pose features. Sequences were also sub-sampled into 20 second segments and Latent Dirichlet Allocation performed to obtain the 25 visual topics that made up each document. Both visual words and topics were used as temporal feature descriptors for each class. All sequences from the CONVERSE set were utilized, with 10 fold cross validation used to evaluate the performance. The RF classifier was produced using 100 trees with random sampling with replacement. The SVM was trained using a radial basis function kernel on the same training set.

It was found that visual topics provide a generalization of the classes which benefit SVM and RF performance (Table \ref{table:cmvt}), while KNN produced more accurate classification on data at the visual words level (Table \ref{table:cmvw}). The importance of each feature was identified via novel use of particle swarm optimization (PSO) to generate a Ranked Feature SVM (SVM-R) classifier, reducing the dimensionality of the feature space and simultaneously performing optimal SVM model selection. The PSO method locates the optimal hyper-parameters that are used to subsequently train the SVM-R classifier by selecting towards correct identification of training samples, removal of redundant features, and the selection of compact feature vectors. This method significantly improved over the previous methods due to the selection of key partitioning features, increasing the accuracy on both visual word and topic feature sets. SVM-R optimization achieved 89.1\% and 87.3\% accuracy for word and topic respective levels of generalization due to its optimized feature set pruning. More detail regarding the use of the SVM-R classifier can be found in \cite{shangming2010,Deng}.

Although these accuracy rates are relatively high, the results have been obtained on subject specific classification utilizing features extracted from long temporal segments of the observation. The main challenge we propose with CONVERSE is for the role of global recognition across multiple subjects for these complex interaction classes.

\begin{table}
\caption{Classification results using visual words (\%).}
\centering
\small{
\begin{tabular}{|l||c|c|c|c|}
\hline
& \multicolumn{4}{c|}{Face\&Pose}\\
\cline{2-5}
& KNN & RF & SVM & SVM-R\\ 
\hline 
Describing Work & 81.2 & 90.6 & 88.4 & 100.0\\
\hline 
Story Telling & 59.7 & 51.0 & 70.6 & 80.2\\
\hline 
Problem Solving & 41.4 & 12.8 & 35.1 & 80.7\\
\hline 
Debate & 55.3 & 51.6 & 67.7 & 91.8\\
\hline 
Discussion & 50.0 & 62.7 & 69.5 & 61.1\\
\hline 
Subjective Question & 30.8 & 5.2 & 35.8 & 91.7\\
\hline 
Jokes & 36.3 & 14.2 & 47.7 & 80.0\\
\hline 
\hline 
Average & 50.7 & 41.2 & 59.3 & 89.1\\
\hline 
\end{tabular} 
}
\label{table:cmvw}
\end{table}

\begin{table}[!ht]
\caption{Classification results using visual topics (\%).}
\centering
\small{
\begin{tabular}{|l||c|c|c|c|}
\hline
& \multicolumn{4}{c|}{Face\&Pose}\\
\cline{2-5}
& KNN & RF & SVM & SVM-R\\ 
\hline 
Describing Work & 63.5 & 91.7 & 76.4 & 100.0 \\
\hline 
Story Telling & 35.1& 73.2 & 68.3 & 80.2\\
\hline 
Problem Solving & 37.1 & 73.6 & 74.3 & 80.7 \\
\hline 
Debate & 48.6 & 73.6 & 67.1 & 81.97 \\
\hline 
Discussion & 38.4 & 78.7 & 63.5 & 61.11 \\
\hline 
Subjective Question & 22.5 & 63.3 & 63.5 & 91.74 \\
\hline 
Jokes & 27.5 & 70.3 & 66.3 & 80.0 \\
\hline 
\hline 
Average & 38.9 & 74.9 & 68.5 & 87.3 \\
\hline 
\end{tabular} 
}
\label{table:cmvt}
\end{table}

\section{Conclussion}
\label{sec:Conc}
This paper presents the current state of the art in regards to the datasets that are available to the \ac{HAR} community, highlighting the need for a dataset that presents subtle interactions between two individuals. The field has progressed over the previous decades, moving from the simplistic single action sequences towards a more natural representation of daily actions and interactions. We also provide clear definitions regarding the level of abstraction within the observations that are commonly encountered in the field, placing our proposed dataset within that of complex conversation interaction rich activities.
By using pose based techniques we have shown that the recognition of top level action classes within the CONVERSE dataset is possible from using pose estimation output obtained from the Kinect sensor, \cite{Deng,Deng2013a,Deng2013k,Deng2013}. We have utilized current techniques, such as the Bag of Key Words, to describe the higher level event in terms of the composition of lower level action primitives. 
The full dataset is made publicly available for further research into the understanding of highly complex interactions at \cite{data_CONVERSE}.

\section*{References}
\bibliographystyle{IEEEtran}
\bibliography{dataset_RefAcr,dataset_Ref}
\end{document}